\pdfoutput=1

\documentclass[11pt]{article}

\usepackage[preprint]{acl}

\usepackage{times}
\usepackage{latexsym}

\usepackage[T1]{fontenc}

\usepackage[utf8]{inputenc}

\usepackage{microtype}

\usepackage{inconsolata}

\usepackage{graphicx}

\usepackage{multirow}

\usepackage{comment}
\usepackage{url}

\title{StandUp4AI: A New Multilingual Dataset for Humor Detection in Stand-up Comedy Videos}

\author{Valentin Barriere\thanks{Same supervision} \\
  Universidad de Chile -- DCC \\
  Santiago, Chile \\
  \texttt{vbarriere@dcc.uchile.cl} \\\And
  Nahuel Gomez \\
  Universidad de Chile -- DIE \\
  Santiago, Chile \\
  \texttt{nahuel.gomez@ug.uchile.cl} \\\And
  Leo Hemamou \\
  Without Affiliation \\
  Paris, France\\
  \texttt{l.hemamou@gmail.com} \\\AND
  Sofia Callejas \\
  INRIA Chile \\
  Santiago, Chile \\
  \texttt{sofia.callejas@inria.cl} \\\And
  Brian Ravenet$^*$ \\
  Université Paris Saclay -- LISN\\
  Orsay, France\\
  \texttt{brian.ravenet@universite-paris-saclay.fr} \\}

\begin{document}
\maketitle 
\begin{abstract}
Aiming towards improving current computational models of humor detection, we propose a new multimodal dataset of stand-up comedies, in seven languages: English, French, Spanish, Italian, Portuguese, Hungarian and Czech. 
Our dataset of more than 330 hours 
is automatically annotated in laughter (from the audience), and the subpart left for model validation is manually annotated.  
Contrary to contemporary approaches, we do not frame the task of humor detection as a binary sequence classification, but as word-level sequence labeling, in order to take into account all the context of the sequence and to capture the continuous joke tagging mechanism typically occurring in natural conversations. 
As par with unimodal baselines results, we propose a method for e propose a method to enhance the automatic laughter detection based on Audio Speech Recognition errors. 
Our code and data are available online: \url{https://tinyurl.com/EMNLPHumourStandUpPublic}
\end{abstract}

\section{Introduction and Related Works}

\begin{figure*}
    \centering
    \includegraphics[width=1.\linewidth]{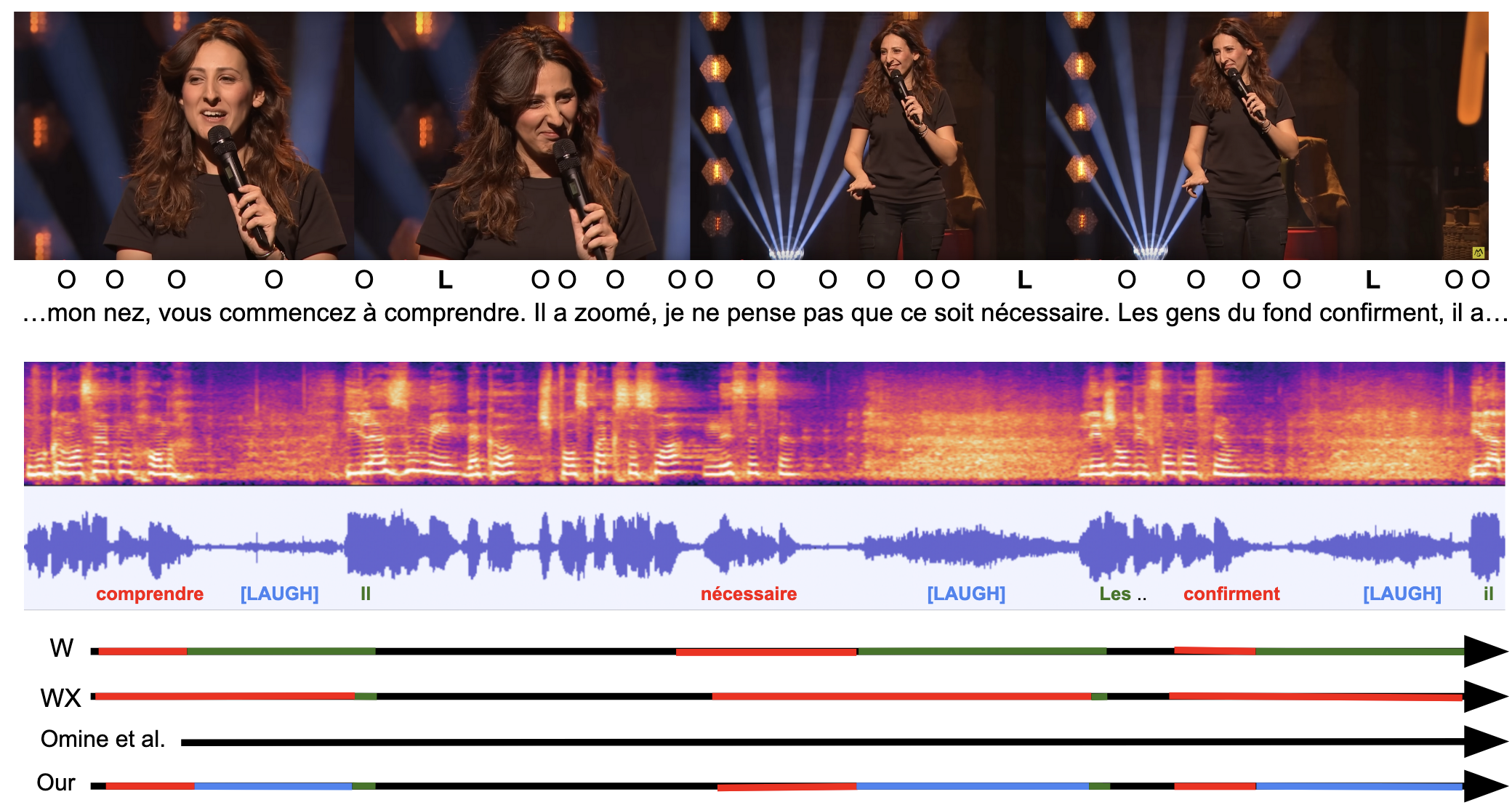}
    \caption{Overview of humor detection modeled as a sequence labeling task, and the method relying on complementary errors from the ASR outputs. 
    \citet{Omine2024} model detected no laughter. Video \href{https://youtu.be/OxvCVuGQ-uk?feature=shared&t=42}{available here}} 
    \label{fig:figure_all}
\end{figure*}

Humor detection 
remains a challenging tasks for computer systems \cite{kalloniatis2024computational,Hyun2024}. 
%
%
Yet, such mechanisms could be a massive improvement, in particular for conversational interactive systems such as chatbots and socially interactive agents. These kind of systems, which are designed to simulate a natural human-like conversation and its structure \cite{Ludusan2022}, often struggle to identify or handle humorous attempts from the user, leading to inefficient and frustrating experiences \cite{zargham2023funny}. 
While different theories of humor 
exist, most of them have in common the idea that humor emerges when the current situation surprisingly deviates from our expectations \cite{warren2016differentiating}. Generally, a joke or funny story is based on the following sequence : the setup of a joke introduces some expectations on how a story usually ends and the punchline reveals the reality and the unexpected (and funny) twist of the story \cite{martin2018psychology}. Sometimes, additional funny comments called \textit{tags} can be added around and after the punchline to maintain the momentum of the laughter. 
Conversational humor often stems from unexpected deviations in content, behavior, or context, with timing and intensity being critical yet unpredictable triggers \cite{wyer1992theory}. Despite theoretical models, comedians rely on live testing to refine timing, phrasing, and delivery for audience engagement \cite{raskin1979semantic}, as responses depend on cultural and contextual factors. This highlights the complexity of modeling humor computationally, necessitating diverse datasets to capture its multifaceted dynamics.
Stand-up comedy, due to its nature aiming at recreating the spontaneity of everyday conversational humor, is a great context for studying these mechanisms and structures with computers. 

Many previous works investigating computational techniques to process humor relied on corpus of people speaking in less natural and more conventional and standardized ways. For instance, in \cite{Purandare2006,Bertero2016a,Patro2021,Liu2024b}, the authors relied on acted data from sitcoms. The UR-FUNNY and Ted Laughter \cite{Hasan2019,Chen2017f} datasets are composed of TED talks, which contain less outbursts of laughter and poorer language diversity than stand-up comedy. Most of the work participating in The MuSE challenges for the automatic estimation of humor are relying on public interviews \cite{Amiriparian2023,Amiriparian2024}, using the Passau-Spontaneous Football Coach Humour dataset \cite{Christ2022}. 
Additionally, while some of the previous work explored other languages \cite{Chauhan2021}, most of them are investigating english humorous content only.
Another early work on stand-up humor is the one of \citet{Turano2022}, which analyzes 90 scripts of 68 comedians, in English only. 
The closest work from ours would be the one described in \citet{Kuznetsova2024a}, which proposed a 40 hours dataset in Russian and English. 

Most humor detection models in videos treat humor as a sequence classification task, identifying punchlines only at the end of a sequence \cite{Choube2020,Hasan2019,Kuznetsova2024a,Liu2024b}. However, multiple laughs can occur within a single sentence, sometimes consecutively. To address this, we reframe the task as sequence labeling, enabling continuous prediction of audience laughter throughout the joke, rather than relying on end-only classification.

In this article, we are presenting multiple contributions towards the development of humor detection models. 
First, we collected and annotated a dataset of stand-up comedy performance in different languages extracted from online videos. This dataset is the largest and most linguistically diverse multilingual dataset of live comedy performances. It has the ambition to be a reference dataset for any type of humor modeling tasks. 
Second, we propose a original methodology for the task of humor detection by using a sequence labeling approach we adapted to automatically predict laughter during a performance. 
Third, this led us to came up with new techniques for handling errors in automatic transcription and automatic laughter detection, validated on a manually laughter annotated test set. 
Fourth, we present first results of sequence labeling models built on our dataset and applied to predict laughter to be used as baselines by the community. 




\begin{table*}[]
    \centering
    \begin{tabular}{l|c|c|c|c|c}
       Youtube Channels  & Language & Videos & Hours & Words & Laughter\\
       \hline \hline
         Comedy Central & \multirow{2}*{English}
 & 263 & 51.2 & 442,904 & 25,772 \\
         Comedy Central UK & & 319 & 18.6 & 174,369 & 13,486\\
         Comedy Central Latam & \multirow{2}*{Spanish} & 971 & 59.2 & 499,329 & 21,708
\\
         Comedy Central España &  & 404 & 18.0 & 150,256 & 5,215
\\
         Comedy Central Italia & Italian & 567 & 55.0 & 433,417 & 12,248
\\
         Comedy Central Magyarország & Hungarian & 73 & 11.4 & 78,002 & 6,875
\\
         Paramount Network CZ & Czech & 123 & 11.5 & 75,806 & 6,129
\\
         Montreux Comedy & French & 652 & 86.0 & 814,727 & 26,789
\\
         Comedy Central Brasil & Portuguese & 245 & 23.4 & 218,592 & 9,972
\\ \hline
          \textbf{Total} & \textbf{MLing} & \textbf{3,617} & \textbf{334.2} & \textbf{2,887,402} & \textbf{128,194}

    \end{tabular}
    \caption{The collection of videos retained for the dataset. Laughter is the number of words labeled as laughter.} \vspace*{-.5cm}
    \label{tab:youtubechannels}
\end{table*}


\section{Dataset}
\label{sec:dataset}

The StandUp4AI dataset is composed of 3,617 standup videos in 7 languages. It contains the associated transcriptions and audience laughters that have been automatically refined, of comedians during Stand-up comedy performances in various languages. To build the dataset, we first collected a specific set of videos of Stand-up comedy from the internet, we then performed automatic transcription on these videos and we finally fixed some errors in the outputs by developing improved transcription and automatic laughter annotation techniques (overview in Figure \ref{fig:figure_all}).

\subsection{Video Recollection}

In total, we gathered 334 hours of video in 7 morphologically diverse languages\footnote{latin, germanic, slavic, and uralic} which is around 3M words and 130k laughter labels. Table \ref{tab:youtubechannels} illustrates the quantity of videos collected per channel and per language. On each channels, we excluded videos from the \textit{Youtube Shorts} section and videos where more than one comedian appeared.

\subsection{Automatic laughter detection}

The next step was to run a task of automatic laughter detection on the videos. The detected laughter would be used to identify and automatically annotate funny events in the performance. We originally based this task on the approach of \citet{Kuznetsova2024a}, who used an off-the-shelf model \cite{Gillick2021}. In our case, we used the state-of-the-art model of \citet{Omine2024}, which has shown better performances for this task. 

\subsection{Transcript extraction}

We perform transcript extraction on each sample using two Audio Speech Recognition (ASR): Whisper \cite{Radford2023} and WhisperX \cite{Bain2023}. These tools allowed us to obtain the timestamped full script of the comedians' performance.


\subsection{ASR-Based Automatic Laughter Detection}

\paragraph{Error Detection}
The timestamps obtained from the ASR were inconsistent for words around events such as laughters and "mouth noises" that activate the ASR's voice activity detection. Such words were frequently assigned an incorrect begin or end timestamps, as the laughter duration would be added to the word duration (and the laughter not detected). 
As this would make the data unreliable to build our model, we engineered a correction by aggregating the outputs of both Whisper and WhisperX. 
%
%
When laughters are perturbating the timestamps of surrouding words, Whisper tends to merge the laughter duration with the next word while WhisperX tends to merge it with the previous one (see Table \ref{tab:error} in Appendix). 
To fix this, we first searched for the 
longer-in-time words, and checked for intersections between both transcripts. Once found, we kept the begin and end timestamps of the intersection to insert a new laughter in the resulting transcript, and we removed the intersection from the previous and next word timestamps. 
This method provides a solution to the problem of erroneous timestamps, and extract potential candidates not discovered by the initial laughter detector.

\paragraph{Automatic Candidate Laughter Validation}
In order to select or not a candidate laughter, we manually annotated the candidates detected in 50 videos with respect to whether or not they were real laughter. We subsequently train a Random Forest classifier on these examples using classical acoustic features. More details are available in Appendix \ref{app:candidate}. 

\begin{table*}
\centering
\begin{tabular}{ll|rrrrrrrrr|r}
 Lang. & Laughters & CS &  EN &   ES &   FR &   HU &   IT &   PT &  Avg. \\ \hline \hline
\multirow{2}{*}{Multiling.} & Raw       & 47.4 & 40.4 & 41.4 & 41.8 & 48.4 & 39.5 & 36.8 & 42.2   \\ 
 & Enhanced & 47.1 & 40.3 & 40.4 & 42.4 & 48.7 & 39.5 & 38.1 & \textbf{42.4}   \\ \hline
 Monoling. & Enhanced  & 41.8 & 38.4 & 37.4 & 42.6 & 45.4 & 35.6 & 34.4 & 39.4 \\ 
\end{tabular} 
\caption{F1 scores by model language and data. Enhanced means trained with laughter from our ASR-based method.} \vspace*{-.5cm}
\label{tab:results}
\end{table*}

\vspace*{-.1cm}
\subsection{Laughter Detection as Sequence Labeling}

We prepare the task of laughter prediction as a sequence labeling task, motivated by the idea that a simple sentence can contains many humorous events that would expect laughter. Each word was labeled with a binary tag indicating whether laughter occurs right after it and before the end of next word. In this way, the model predicts in advance if there will be a laughter event. Further details are provided in Appendix \ref{app:laughters_seq_lab}.

\subsection{Test Set Annotation} \vspace*{-.1cm}

Following the protocol of \citet{Kuznetsova2024a}, we manually annotate a test set composed of 70 videos (10 per language). These samples have been manually annotated in laughs with precise timestamps at 0.1 seconds, using the audio file and audacity. 
The ASR outputs have been manually checked to ensure that the labels are true. 
The test set is used to validate both the laughter detection method based on the ASR outputs and 
acoustic classifier, and the sequence labeling models. 

\vspace*{-.1cm}
\subsection{Other Features}
Even though we did not add them in the current baseline experiments, we also release a set of features we extracted from the data. We extracted Action Units using the \texttt{LibreFace} library \cite{Chang2024a}, poses using the \texttt{MMdetection} and \texttt{MMpose} libraries \cite{mmdetection,mmpose2020}, and the camera angle changes using \texttt{PySceneDetect} library \cite{Castellano2014}. 
We release these features with the data and plan to investigate how they can contribute to the task in future works. 

\vspace*{-.2cm}
\section{Experiments and Results} \vspace*{-.1cm}
\label{sec:experiments}
We conducted two types of experiments. The first validate the proposed technique to find new outbursts of laughter that were not detected using the off-the-shelf model of \citet{Omine2024}. The second is the sequence labeling task. 

\vspace*{-.1cm}
\subsection{ASR-based Laughter Detection Validation} 

Using the proposed method, we obtained 376 outburst candidates on 50 videos that were manually annotated into real laughter or other event. 208 of them were real outbursts of laughter non previously detected by the off-the-shelf laughter detection model. 
We extracted acoustic features with librosa \cite{brian_mcfee-proc-scipy-2015} and trained a random forests to binary detect an outburst. Results are shown in Table \ref{tab:candidate}. The overall method allows detecting approximately 3 new outbursts of laughter per video. More details on the features and models are available in Appendix \ref{app:candidate}.

\begin{table}[h!] \vspace*{-.1cm}
\centering
\begin{tabular}{l|c|c|c}
 & Prec. & Rec. & F1 \\
\hline \hline
Other & 0.79 & 0.87 & 0.82 \\
Laughter & 0.89 & 0.81 & 0.85 \\
\hline
Macro & 0.84 & 0.84 & 0.83 \\
\end{tabular}
\caption{Performances of the Random Forest Candidate Laughter Classifier} \label{tab:candidate} \vspace*{-.3cm}
\end{table}

We validate the whole laughter detection system on the manually annotated test set task with the Intersection over Union (IoU). With a IoU threshold of 0.2,\footnote{if IoU > 0.2, prediction is considered as positive} we obtained a F1-score of 0.51 using the off-the-shelf model, versus a F1-score of 0.58 using our method. More details in Appendix \ref{app:iou}.

\subsection{Sequence Labeler}

We trained unimodal pretrained transformer models \cite{Lample2019} based on the text input in order to predict laughter at the word-level in a binary way. A maximum sequence length of 512 was used with a stripe of 128 when cutting from the same monologue to ensure past context. We optimized it with Adam \cite{Kingma2014}, 10 epochs, and a learning rate of $1e-5$. 
We validate the models with classification metrics and not rigid sequence classification metrics such as seqeval \cite{seqeval} because of the task difficulty. 

\paragraph{Experimental Protocol}
The \texttt{transformers} library \cite{Wolf2019} was used to access the pre-trained \texttt{xlm-roberta-base} and to fine-tune sequence labeling models. The random forests were trained using scikit-learn \cite{Pedregosa2012}.
Experiments were run using torch 2.1.2 \cite{Abadi2016}, transformers 4.46.3 \cite{Wolf2019}, a GPU Nvidia RTX-A6000 and CUDA 12.2. 

\paragraph{Results}

Results are shown in Table \ref{tab:results}. 
First, the multilingual models trained with the raw outputs obtained from \citet{Omine2024}'s laughter detection (Raw) and the ASR-based one (Enhanced) are compared. Results show that the models trained on the cleaned data are reaching higher performances, indicating a second time the quality of ti, as the proposed treatment helps to enhance the quality as the data as training material. 
Second, the results of the multilingual model are compared with the ones of the monolingual models, highlighting the interest of the diversity of our corpus.

\vspace*{-.1cm}
\section{Conclusion}\vspace*{-.1cm}
\label{sec:conclusion}

In this article we presented the most diverse dataset of multilingual stand-up comedy performance at the date of today, StandUp4AI. We propose baseline results on the tasks of laughter prediction approached as a sequence labeling task, highlighting the interest of the diversity contained in our dataset. 
On top of this, we show the interest of a simple yet efficient technique enhancing a state-of-the-art automatic laughter detection method, that we successfully validate with manual annotations and by using it to train a humor detection model. 
The results highlighted the potential of our dataset for the development of computational models of humor.

\section{Limitations}

This work faces several limitations. First the humor detection task only focuses on unimodal textual model for now. This is by design as we decided to focus on unimodal approach in order to acquire initial results before moving towards multimodal models in future works.

Second, we do not take into account different intensity of the laughter. This is because there is a significant variability in acoustic intensity in the collected videos. We plan to address this, by performing at least a normalization, and to include this additional dimension in future steps. 

Third, a more thorough analysis of the ASR errors would be beneficial. Dialect languages such as Mexican or Chilean Spanish can be challenging for the speech-to-text models, especially for discourses where slang and vulgarity play a big part. However, we believe that this is a small portion of the whole dataset and does not impact its global quality. 

Finally, the paper relies on Youtube Videos that can be subject to deletion. However, we do not release neither the video nor audio content, just the metadata and annotations, like other famous corpora \cite{Zadeh2020,Zadeh2019socialiq,Hasan2019}. 


\bibliography{custom,humour_standup,JRC}

\appendix


\begin{table*}
\centering
\begin{tabular}{llrrrrrrrr|r}
 Lang. &   CS &  EN &   ES &   FR &   HU &   IT &   PT &  Avg \\ \hline \hline
Monoling. Raw & 34.8 &   26.5 & 24.5 & 27.6 & 40.5 & 15.0 & 19.6 &   26.9 \\
 Monoling. Enhanced & 36.4 &   27.4 & 27.3 & 27.2 & 40.1 & 14.2 & 22.7 &    27.9 \\
Multiling. Enhanced & 35.8 &   27.5 & 27.4 & 28.0 & 42.8 & 17.5 & 21.9 &  \textbf{28.7} \\
\end{tabular}
\caption{F1 scores by model type and test language, using the \textbf{automatic} laughter detection. Enhanced means model trained on our ASR-enhanced set of laughters.} \vspace*{-.5cm}
\label{tab:results_asrbased}
\end{table*}

\section{Labels Creation} \label{app:laughters_seq_lab}

Every word was tagged so that its label means that the agent should laugh right after it, or that is should continue to laugh. With this method, the agent can predict when to start and stop laughing, before it actually happens to the audience. 
For each laughter segment with start $t_0$ and end $t_1$, we first locate the “start” word by finding the word whose timing window either overlaps or immediately follows the laughter’s $t_0$; similarly we find the “end” word around $t_1$. If both boundaries fall on the same word, that word is labeled positive. Otherwise, all the words in between are tagged as positive. 

%

\section{Candidate Laughter Selection} \label{app:candidate}

\paragraph{Acoustic Features}
The acoustic features extracted can be categorized into several groups, including temporal characteristics such as duration, voiced ratio, voiced frames, burst count, and temporal centroid. Additionally, features related to energy and amplitude were extracted, including rms mean, rms standard deviation, rms slope, energy at the 90th percentile, and root mean square (rms). Spectral features were also considered, comprising spectral bandwidth, spectral rolloff at 85\% and 95\%, spectral flatness, spectral contrast, and spectral centroid. Furthermore, pitch-related features such as pitch median, pitch standard deviation, and harmonics-to-noise ratio (hnr) were included, along with modulation energy between 4-12 Hz. The feature set was further enriched with chroma features (chroma 1-12), Mel-frequency cepstral coefficients (mfcc 1-13), and their first and second derivatives (delta mfcc 1-13 and delta2 mfcc 1-13), providing a detailed representation of the audio signals' spectral and temporal properties.

\paragraph{Classifier}

Once the acoustic features were extracted and verified, a two-stage pipeline was implemented: \textbf{verification} and \textbf{classification}.

In the verification stage, all audio segments with a duration shorter than 0.5 seconds were discarded and classified as \textit{‘other’}. These cases were not considered in the evaluation of the model’s performance.

Subsequently, a \textit{Random Forest} classifier was applied. For hyperparameter tuning, 15\% of the dataset was randomly selected, focusing on the parameters \texttt{n\_estimators}, \texttt{max\_depth}, and \texttt{min\_samples\_split}, whose optimal values were 50, 13, and 2, respectively.

With the selected hyperparameters, 200 iterations were performed, varying the training/testing split in each run, while consistently using 15\% of the data for validation.

The classifier was designed as a binary model, distinguishing between the \textit{laughter} and \textit{non-laughter} classes. The latter included events such as fillers, claps, silence, and general noise.

The 95\% confidence intervals for the performance metrics obtained were as follows:

\begin{table}[h!] \vspace*{-.1cm}
\centering
\begin{tabular}{l|c|c|c}
\textbf{Class} & \textbf{Precision} & \textbf{Recall} & \textbf{F1 Score} \\
\hline \hline
Non-laughter & 0.77--0.80 & 0.85--0.88 & 0.81--0.83 \\
Laughter     & 0.88--0.90 & 0.80--0.82 & 0.83--0.85 \\
\hline
\textbf{Macro avg.} & 0.83--0.85 & 0.83--0.85 & 0.82--0.84 \\
\end{tabular}
\caption{95\% confidence intervals for the performance metrics of the Random Forest laughter classifier}
\label{tab:rf_95_confidence}
\vspace*{-.3cm}
\end{table}

\section{Correcting Timestamp Errors} \label{app:timestamps}

Table \ref{tab:error} shows the principal of the algorithm we used to correct the timestamps errors of WhisperX and Whisper around outbursts of laughter.  

\begin{table}[!h]
    \centering
    \begin{tabular}{llcr}
        & Word1 & [Laugh] & Word2  \\ \hline \hline
        WhisperX & $t_{0,w_1}$ & \multicolumn{1}{r}{$t_{1,w_1}$} & $t_{0,w_2}$ $t_{1,w_2}$ \\
        Whisper & $t'_{0,w_1}$ $t'_{1,w_1}$ & \multicolumn{1}{l}{$t'_{0,w_2}$} & $t'_{1,w_2}$ \\ \hline 
        Our & $t'_{0,w_1}$ $t'_{1,w_1}$ & $t'_{0,w_2}$ $t_{1,w_1}$ & $t_{0,w_2}$ $t_{1,w_2}$ \\
    \end{tabular}
    \caption{Example of errors in the ASR outputs}
    \label{tab:error}
\end{table}


\section{ASR-based Acoustic Laughter Detection} \label{app:iou}

We validate the ASR-based Acoustic Laughter Detection method on the manually annotated test set. We used the Intersection over Union to validate the quality of the predictions. Using a threshold of 0.2, we obtained the results in Table \ref{tab:iou}.

\begin{table}[h!]
\centering
\begin{tabular}{l|c|c|c}
 & Prec. & Rec. & F1 \\
\hline \hline
\citealt{Omine2024} & 0.68 & 0.41 & 0.51 \\
All Candidates & 0.62 & 0.52 & 0.56 \\
Filtered (RF) & 0.70 & 0.49 & 0.58 \\
\end{tabular}
\caption{Performances of the ASR-based Acoustic Laughter Detection methods on the Manually Annotated Test Set} \label{tab:iou}
\end{table}

\section{Humor Detection on the Automatic Test Set}

The performances of the model on the test set, when using automatic laughter detection (not manual) are shown in Table \ref{tab:results_asrbased}. The performances are 14 points lower than when comparing with the ground truth. This means that, even though trained with weak labels, the system achieves to detect the real humor case.

\end{document}